\documentclass[sigconf]{acmart}
\usepackage[utf8]{inputenc}

\newcommand{\sys}{{Student Parallelism}\xspace}

\usepackage[normalem]{ulem}
\usepackage{algorithm}
\usepackage{algpseudocode}
\usepackage{mathtools}
\usepackage{comment}
\usepackage{subfigure}
\usepackage{xspace}
\usepackage{color}
\usepackage{multirow}
\usepackage{makecell}
\usepackage{graphicx}
\usepackage{enumitem}

\usepackage{pifont}

\begin{document}

\title{
Exploiting Student Parallelism for Efficient GPU Inference of BERT-like Models in Online Services
}

\settopmatter{printacmref=false} 
\renewcommand\footnotetextcopyrightpermission[1]{} 
\pagestyle{plain} 

\hyphenation{Aca-de-mus}

\begin{abstract}
Due to high accuracy, BERT-like models have been widely adopted by text mining and web searching.
However, large BERT-like models suffer from inefficient \emph{online} inference,
facing the following two problems on GPUs:
(1) their high accuracy relies on the large model depth, 
which linearly increases the sequential computation on GPUs;
(2) stochastic and dynamic online workloads cause extra costs from batching and paddings.

Therefore, we present \sys for the real-world setting of GPU inference on online workloads.
At its core, \sys adopts stacking distillation and boosting ensemble, 
distilling the original deep model 
into a group of shallow but virtually stacked student models running in parallel.
This enables \sys to achieve a lower model depth (e.g., two layers) than the others
and the lowest inference latency while maintaining accuracy.
In addition, adaptive student pruning realizes dynamic student numbers according to changing online workloads.
Especially for occasional workload bursts, 
it can temporarily decrease the student number with minimal accuracy loss to improve system throughput. 
We conduct comprehensive experiments to verify the effectiveness,
whose results show that \sys outperforms the baselines
by $4.1\times\sim 1.6\times$ in latency while maintaining accuracy 
and achieves up to $22.27\times$ higher throughput for workload bursts. 
\end{abstract}

\author{Weiyan Wang}
\affiliation{%
  \institution{Tencent \\
  Hong Kong University of Science and Technology}
  \country{China}
}

\author{Yilun Jin}
\affiliation{%
  \institution{
  Hong Kong University of Science and Technology}
  \country{China}
}
\author{Yiming Zhang}
\affiliation{%
  \institution{
  Xiamen University}
  \country{China}
}
\author{Victor Junqiu Wei}
\affiliation{%
  \institution{
  Macau University of Science and Technology}
  \country{China}
}
\author{Han Tian}
\affiliation{%
  \institution{
  Hong Kong University of Science and Technology
  }
  \country{China}
}
\author{Li Chen}
\affiliation{%
  \institution{
  Zhongguancun Lab.
  }
  \country{China}
}
\author{Jinbao Xue}
\affiliation{%
  \institution{
  Tencent
  }
  \country{China}
}
\author{Yangyu Tao}
\affiliation{%
  \institution{
  Tencent
  }
  \country{China}
}
\author{Di Wang}
\affiliation{%
  \institution{
  Tencent
  }
  \country{China}
}
\author{Kai Chen}
\affiliation{%
  \institution{
  Hong Kong University of Science and Technology
  }
  \country{China}
}
\maketitle
\section{Introduction}

Due to the marvelous success in natural language understanding,  
BERT-like models~\cite{devlin2018bert,liu2019roberta,he2021deberta} have been the foundation model for discriminative tasks like text mining and web searching in online services.
However, large BERT-like models pay the price of high inference time for the model capability.
Besides high accuracy, a wide range of online services are time-sensitive, requiring efficient \emph{online inference}. 
For example, news sentiment classification for automatic stock trading should be quicker than any other, 
because even a sub-millisecond delay can determine the profit~\cite{aquilina2022quantifying};
for good user experiences, Retrieval Augmented Generation(RAG)~\cite{lewis2020retrieval} usually has a ten-millisecond-level time budget~\cite{yang2020model} for every query to search the relevant web pages as the reference of answer generation;
and online writing assistants need to give immediate responses to text streams while users are typing~\cite{DBLP:conf/acl/KanekoMKSI20}. 

\begin{figure}[t]
	\centering
	\subfigure[Model Depth (Layer \#)]{
		\begin{minipage}[b]{0.45\linewidth}
			\includegraphics[width=1\linewidth]{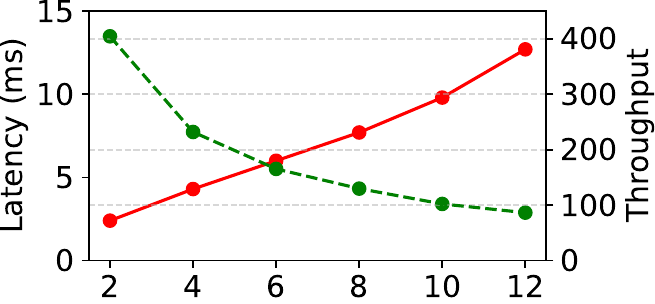} 
		\end{minipage}
		\label{fig:model_depth}
	}
	\subfigure[Model Width (Hidden Dim.)]{
		\begin{minipage}[b]{0.45\linewidth}
			\includegraphics[width=1\linewidth]{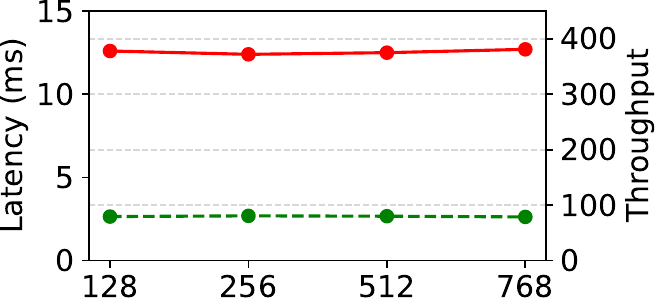} 
		\end{minipage}
		\label{fig:model_width}
	}
    \subfigure[Input Batch Size]{
    		\begin{minipage}[b]{0.45\linewidth}
   		 	\includegraphics[width=1\linewidth]{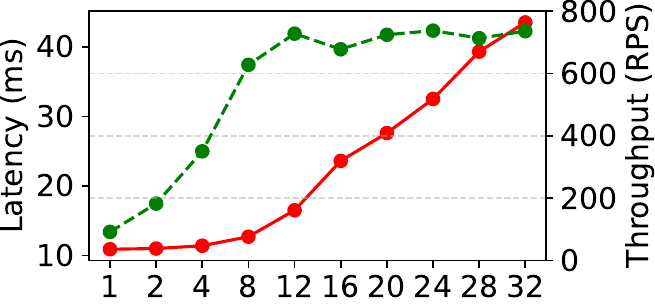}
    		\end{minipage}
		\label{fig:batch_size}
    }   	
	\subfigure[Input Sequence Length]{
		\begin{minipage}[b]{0.45\linewidth}
			\includegraphics[width=1\linewidth]{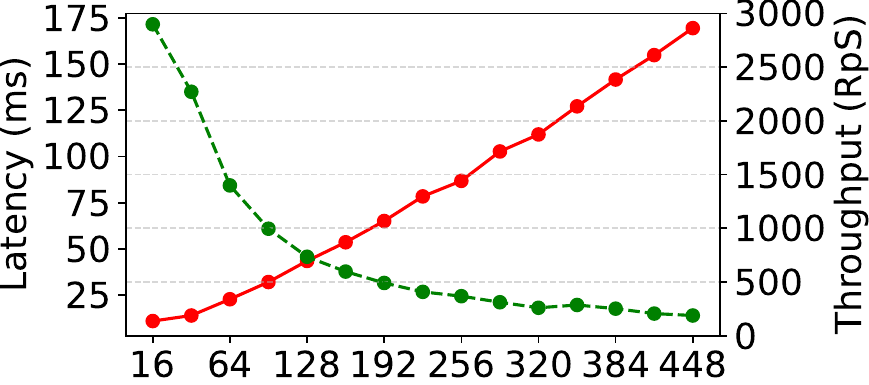} 
		\end{minipage}
		\label{fig:seq_len}
	}
	\caption{ Influences of different model and input factors on latency in red solid line) and throughput (green dashed line) }
	\label{fig:input_factors}
	\vspace{-8mm}
\end{figure}

Previous efficient inference works 
such as compact model design~\cite{sun2020mobilebert,xu2021bert}, knowledge distillation~\cite{sanh2019distilbert, sun2019patient, jiao2019tinybert}, and pruning~\cite{DBLP:conf/rep4nlp/GordonDA20,michel2019sixteen, hou2020dynabert}
focus on reducing the model size for weak edge devices like cell phones. 
However, since deep learning essentially relies on model depth for high model capability~\cite{lecun2015deep}, 
these methods prefer reducing model width to model depth (about half the layers remained) for better accuracy.
Moreover, early existing methods~\cite{DBLP:conf/acl/XinTLYL20, DBLP:conf/acl/LiuZWZDJ20,zhou2020bert} still have some hard samples to go through all layers. 
Therefore, their deep architectures largely degrade their inference performance on parallel computing GPUs which are the de-facto device for online inference services.  
Figure~\ref{fig:model_depth} shows that model depth linearly increases the latency due to serial forward propagation.
When GPU has enough parallel computation capability, it keeps the same latency and throughput for different model widths as shown in Figure~\ref{fig:model_width}.
Therefore, these deep and thin models are not suitable for efficient GPU inference. 

Furthermore, existing works overlook the opportunity to optimize stochastic online workloads, resulting in the extra costs of batching and padding.
Different from collected training data batches, 
online inference samples of different lengths arrive in stochastic and dynamic patterns. 
GPUs need large batches to achieve high throughput via data parallelism~\cite{kwon2023efficient}, 
as Figure~\ref{fig:batch_size} shows that latency remains stable and throughput keeps increasing until the batch size is large enough.
Therefore, it has to employ a waiting queue to batch enough data samples from stochastic online workloads~\cite{zhang2019mark,ali2020batch,DBLP:conf/usenix/Cui00WLZ0G22}, 
and early-arrived samples have higher latency due to waiting. 
Text sequences have different lengths ranging from 10 to hundreds,
but GPUs require all samples in one batch to have the regular length.
Existing works commonly pad short samples into the maximum length to form the regular batch~\cite{devlin2018bert}.
As shown in Figure~\ref{fig:seq_len}, latency can increase rapidly with sequence length due to $O(N^2)$ complexity of self-attention~\cite{DBLP:conf/nips/ZaheerGDAAOPRWY20},
so paddings also bring in significant wasted computations.


Therefore, 
we present \sys to explore \emph{student parallelism} for efficient GPU inference of BERT-like models, 
especially for stochastic online workloads in practice. 
At its core,
\sys distills the original deep model
into an adaptive group of parallel and shallow student models, 
which are virtually stacked.
Therefore, the sequential layer computations can be replaced with
the equivalent parallel computations of students while maintaining accuracy.
To achieve this,
we propose a novel knowledge distillation method 
based on stacking distillation and explicit boosting ensemble(\S~\ref{sec:training}).
Despite the parallel inference, every student is sequentially trained  
to gradually reduce the error 
between the teacher and the group of students.
Every student distills all previous students into its intermediate layer by stacking distillation,
so that it is virtually stacked on all the previous students.
Then all students can make better corrections in the final layer with the estimated output of previous students.
Consequently, all students can maintain accuracy with no more than one-quarter of the original layers (even only two layers sometimes, which is minimal for nonlinear classification),
which significantly reduces the inference latency on GPUs.
Beyond the distillation, it further conducts adaptive student pruning to improve generalization and reduce the student number.  
By dynamically dropping some last-trained students,
\sys improves the generalization of the remained students, 
while pruning the redundant students causing over-fitting.
More importantly, it realizes dynamically adjustable student numbers according to the changing workload.
In case of workload bursts,
it can temporarily reduce the student number to a small value(e.g., one or two) with minimal accuracy loss. 

By leveraging the proposed student parallelism, we further make specialized designs to reduce waiting and padding from data parallelism. 
It allocates the students of the same group on different GPUs to better parallelize even the single sample.   
Then it replicates the student group into multiple copies to run concurrently and share the same set of GPUs. 
By employing the length-aware buffer, it enables immediate and concurrent inference of different samples to be free of batching large and regular data. 

\begin{figure}
    \centering
    \includegraphics[width=\linewidth]{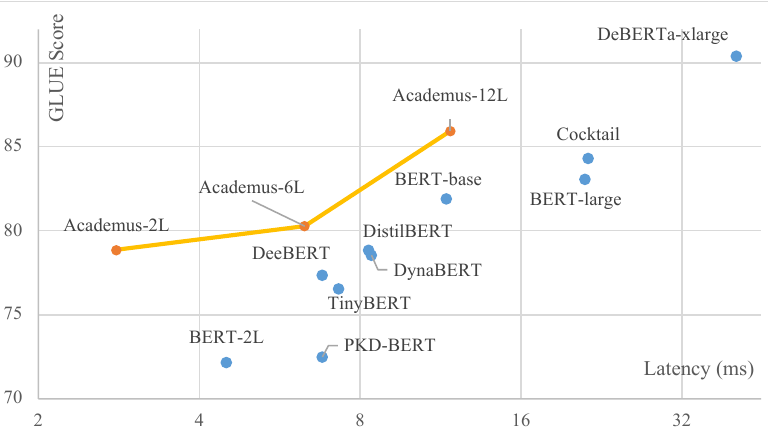}
    \caption{Comparison in the trade-off between accuracy and latency (the upper left corner is optimal)}
    \label{fig:tradeoff}
    \vspace{-8mm}
\end{figure}

Evaluation results (\S~\ref{sec:evaluation}) on real online workload traces show that 
\sys outperforms all baselines by $4.1\times\sim 1.6\times$
in terms of average and tail latency,
while achieving the best prediction accuracy in all baselines without full depth or width.
For workload bursts, 
\sys achieves up to $22.27\times$ improvement in throughput with competitive accuracy
by adaptively reducing the student number.
In addition, our work can effectively compress 24-layer and 48-layer BERT-like models into a quarter of model depth (6 layers and 12 layers respectively), 
while other baselines fail. 
Figure~\ref{fig:tradeoff} demonstrates \sys achieves the best tradeoff between accuracy and latency, in which \sys achieves better accuracy with the same latency budget or lower inference latency when reaching a similar accuracy. 

As far as we know, \sys is the first to consider the real-world setting of efficient GPU inference on online workloads from the view of model design. 
Our technique contributions can be summarized as follows: 
\begin{enumerate}[leftmargin=10pt]
\item  We leverage virtual stacking and boosting ensemble to  distill the deep model into a group of parallel students 
	in the lowest depth for low latency with comparable accuracy. 
\item
 To handle online workloads, we can reduce the student number with minimal accuracy loss for workload bursts, and we also make specialized designs for student parallelism to reduce waiting and paddings.
\item Comprehensive experiments with real workload traces verify the effectiveness of our \sys, 
	where it outperforms other baselines in terms of accuracy, latency, and throughput. 
\end{enumerate}

\section{Related Work}
\label{sec:relatedwork}

Previous works have explored various efficient inference methods for BERT-like models. 
However, they still suffer from high latency due to large model depth
and extra costs for online workloads. 
There are three categories of existing works as follows:
\begin{itemize}[leftmargin=10pt]
\item
\textbf{Model Compression}:
these works only generate a single small model for weak edge devices, 
which still suffer from considerable model depth and high latency on GPUs.
MobileBERT~\cite{sun2020mobilebert} and NAS-BERT~\cite{xu2021bert} make \textit{compact model design} that can achieve the similar accuracy.
\textit{Knowledge distillation} transfers the knowledge from a large teacher~\cite{sanh2019distilbert, sun2019patient, jiao2019tinybert} or teacher ensemble~\cite{asif2020ensemble} to a small student model, so that the small student mimics the behavior of its teacher.
\textit{Pruning} can remove redundant weights, neurons, attention heads or tokens in well-trained models for less costs and better generalization~\cite{hou2020dynabert,malach2020proving,kim2022learned}.  
\textit{Early exiting}~\cite{DBLP:conf/acl/XinTLYL20, DBLP:conf/acl/LiuZWZDJ20,zhou2020bert} adds multiple classifiers on different layers as the multiple exits, so that it allows early exit from bottom layers for "easy" samples.
\textit{Quantization}~\cite{kim2021bert, shen2020q, zafrir2019q8bert, tang2022mkq} replaces slow float operators with fast integer operators, 
which releases higher parallel computation capability.
However, deep learning essentially relies more on model depth than model width for high model capability~\cite{lecun2015deep},
so they still require a considerable large model depth to maintain the accuracy.  
For example, MobileBERT and NAS-BERT are still deep (12 or even 24 layers) but much thinner to avoid accuracy loss.
6-layer Distilbert can maintain 96.3\% overall score of the 12-layer Bert teacher, 
but the 3-layer PKD-BERT can only get 88.5\% (\S~\ref{sec:accuracy}). 
Model pruning~\cite{hou2020dynabert} can only reduce about the half of layers at most to maintain good accuracy. 
In early exiting, the "difficult" samples still have to go through all layers.
Quantization not only needs hardware support in low-bit integers~\cite{zafrir2019q8bert} but also fails to reduce the model depth and computation complexity.
It also relies on large batch sizes to increase parallelism degree and speed-up (e.g., the speedup is $2.4\times$ and $3.3\times$ when the batch size is 1 and 8 respectively~\cite{kim2021bert}). 

\item 
\textbf{Model Ensemble}:
BERT stacks all attention layers with residual connections to enable the multi-path forward,
which can be interpreted as the implicit boosting ensemble of all layers within the deep model~\cite{veit2016residual,huang2018learning}.
The \textit{Mixture-of-Experts (MoE)} in attention heads~\cite{DBLP:conf/iclr/ShazeerMMDLHD17,DBLP:conf/naacl/ZuoZLHZC22} has a router to activate only one or two experts in every layer.
It can keep constant inference costs no matter how wide the model is, 
but it cannot decrease the model depth.
\textit{Bagging ensemble} like Cocktail~\cite{DBLP:conf/nsdi/CrankshawWZFGS17,276950} can select different relatively small models from the above-mentioned models to run in parallel for lower latency and high accuracy.
However, they rely on the different model architectures for the ensemble diversity, resulting in the straggler problem on GPUs. 
\textit{SenseAI}~\cite{wang2021sensai} divides the multi-class model into several pruned binary models,
which still have a large model depth and irregular pruned architectures.
It cannot be used in regression and binary classification tasks either.
\textit{Progressive ensemble distillation}~\cite{dennis2023progressive} can get closer to the teacher’s accuracy by adding more students with different networks. 
However, the maximum layer number of heterogeneous students remains to be large,
and its accuracy decreases significantly if any students are dropped to speed up.
Some \textit{Collaborative and Self-distillation}~\cite{zhu2018knowledge,DBLP:conf/icann/ZainJZ22, sun2021collaborative} explore distilling multiple teachers and existing students into a new better student for higher accuracy, 
but they do not leverage multiple students together for lower model depth.

\item \textbf{Online Workloads Handling}: Existing works in model design usually make the ideal assumption and ignore the fact that online workload is stochastic, 
so that online inference system has to pay extra costs of batching and padding in practice. 
They reserve some over-provisioned GPUs in advance for potential workload bursts, 
which are underutilized for the most time~\cite{zhang2019mark,DBLP:conf/usenix/Romero0YK21,276950}.
According to workload prediction and profiling,  previous system works can dynamically set the batch size and maximum waiting time to reduce the idle waiting~\cite{zhang2019mark,ali2020batch}, 
but it can hardly be optimal. 
Some recent works~\cite{triton,pytorch2018pytorch,abadi2016tensorflow,zhai2022bytetransformer} bring in some operators supporting ragged batches to avoid extra paddings, 
but they can hardly support and perfectly optimize all operators and situations~\cite{DBLP:conf/mlsys/FegadeCGM22}. 

\end{itemize}

In contrast, \sys trains a group of parallel students having the lowest model depth for GPUs, other than a single student for weak edge devices. 
Because of virtual stacking and boosting ensemble, it can even effectively reduce the model depth of large models like BERT-large and DeBERTa-xlarge,
while the others fail.
All students also have the same shallow architecture to be free of stragglers.
Through adaptive student pruning, it can reduce the student number with minimal accuracy loss for higher throughput, being free of over-provisioned GPUs.
Due to student parallelism, it also saves the costs of waiting queues and padding for the stochastic online workload.
Being orthogonal to quantization, it can quantize all students for further improvement.

\section{Methodology}	
In this section, we first show how to combine virtual stacking and boost ensemble to train the student group to generate final representations similar to deep teachers.
Then we discuss how to apply adaptive student pruning in training the final classifier, which achieves better generalization and dynamical student numbers.
Finally, we describe how the multiple students run in parallel for online inference.

\subsection{Virtual Stacking and Boosting Ensemble}
\label{sec:training}

\begin{figure}[t]
\centering
\includegraphics[width = 0.99\linewidth]{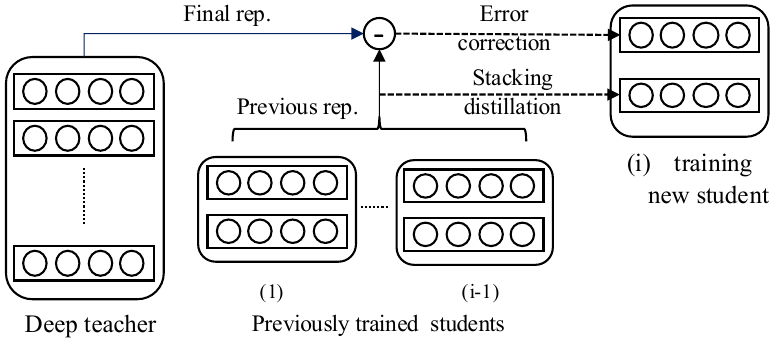}
\caption{Virtually stacked students: the intermediate layer imitates all the previous students and the top layer reduces the residual error. } 
\label{fig:offline_training}
\vspace{-7.5mm}
\end{figure}

First of all, we recall that transformer models like BERT stack all the attention layers with residual connections. 
The residual connection adds the original input of the $i-th$ layer $x_{i}$ back to the result of its layer function $F_i(x_i)$, 
so its output representation is $H_{i}=F_i(x_i)+x_i$, namely the refinement of its input. 
Because there are numerous attention layers stacked with residual connections. 
Then the output of the n-th layer is expanded as $H_n=F_n(\sum_{i=1}^{n-1}H_i+x_0)+\sum_{i=1}^{n-1}H_i+x_0$.
Thus, residual connections enable the multi-path forward in the vision transformer, 
which behaves like the ensemble of networks in different depth~\cite{veit2016residual,huang2018learning}.  

Inspired by the implicit boosting ensemble within BERT, 
we propose to combine virtual stacking and boosting ensemble to trade more parallel student models for less sequential layers, while generating similar final representations of deep teachers.
To be friendly to GPUs, all students share the same shallow model architecture, resulting in weak model capability and homogeneity.
To improve model capability and diversities,
every student is not only added to gradually reduce the error but also virtually stacked via stacking distillation to imitate the direct refinement of its own previous outputs, as illustrated in Figure~\ref{fig:offline_training}.
Overall, the training strategy as described in Algorithm~\ref{algo:boosting} improves diversities from other aspects, including different training loss, different parameter initializations, and different data subsets.



\textit{Boosting Ensemble.}
All students are sequentially trained in the style of boosting ensemble to gradually correct its own residual error on the final representation. 
Except that the first student directly mimics the final representation of the teacher, 
the $i-th$ student learns to correct the error of the ensemble of its own previous students.
We keep adding new students until the overfitting happens on spared validation data (line 3 in Algorithm~\ref{algo:boosting}).
Formally, we define the boosting ensemble of the $M$ students having $N$ attention layers to generate the final representation as follows:
\begin{equation}
	B^{(M)}(x):=\sum_{m=0}^{M} \alpha^{(m)} S^{(m)}_{N}(x)	
\end{equation} 
where $S^{(m)}_{N}$ is the final representation of m-th student model in its $N$-th layer, and $\alpha^{(m)}$ is its multiplier .  
Then we can optimize the following residual training loss $L_{\text{boost}}^{(i)}$ for the i-th student :
\begin{equation}
	L_{\text{boost}}^{(i)}(x)= \frac{1}{2}
	\left\| T(x)-B^{(i-1)}(x)-S^{(i)}_{N}(x) \right\|^2  .
\label{equ:boost}
\end{equation}	
The $T(x)$ stands for the final representation output of the teacher model and $B^{(i-1)}(x)$ is the boosting ensemble of its all previous $i-1$ students. 
And only the i-th training student $S^{(i)}$ is trainable, while the other models like $T$ and $B^{(i-1)}(x)$ are fixed. 
After training the student $S^{(i)}$, we can set its multiplier $\alpha_i$ by line search to minimize the loss $\frac{1}{2}(T(x)-B^{(i)}(x))^2$, 
except that $\alpha_0$ is always 1.

\begin{algorithm}[b]
\vspace{-1mm}
\caption{Sequential Student Training}
\begin{algorithmic}[1]
\Require TeacherModel, PreMiniModel
\State \textbf{\#Sequential training of student models}
\State studentList = [PreMiniModel]
\While {not overfitting}
	\State newStudent = studentList[-1].copy()
	\State Sample the new training dataset
	\While {training}
		\State $t\_rep_{N}$= TeacherModel(x)
            \State $prev\_rep=\sum_j^{k}\alpha_j studentList[j](batch)$
		\State $s\_rep_{N}$, $s\_rep_{\lceil N/2 \rceil}$ = newStudent(x)
		\State Compute $L_{\text{boost}}$ as Equation~\ref{equ:boost} and $L_{\text{stack}}$ as Equation~\ref{equ:stack}
		\State Update the newStudent
	\EndWhile
	\State studentList.append(newStudent)
\EndWhile
\State studentList = studentList[:-1]
\vspace{-1mm}
\end{algorithmic}
\label{algo:boosting}
\end{algorithm}

\textit{Virtual Stacking.} Despite that all students run independently, we propose stacking distillation to make every student virtually stacked on its own previous students.
Any student learns to distill the ensemble of its all previous students into the representation of its $\lceil N/2 \rceil$-th layer. 
Since stacking distillation makes the intermediate representation imitates all the previous students, 
the following layers can better correct the error by refining the intermediate representation, in a similar way to a deep BERT. 
The loss $L_{\text{stack}}^{(i)}$ of stacking distillation for the i-th student is defined as follows:
	\begin{equation}
		L_{\text{stack}}^{(i)}(x) = \frac{1}{2} \left\| B^{(i-1)}(x)-S^{(i)}_{\lceil N/2 \rceil}(x) \right\|^2
	\label{equ:stack}
	\end{equation}
	where $S^{(i)}_{\lceil N/2 \rceil}(x)$ denotes the intermediate representation of the $N/2$-th layer in the i-th student model. In Equation~\ref{equ:stack}, only $S^{(i)}$ is trainable while the other models are fixed.
	
Therefore, the overall training loss $L$ for the whole representation distillation can be written as follows:
	\begin{equation}
		L=L_{\text{boost}}+\lambda L_{\text{stack}}
	\end{equation}
where $\lambda$ is the hyper-parameter to balance the final representation distillation and stacking distillation. 
Since every student is trained under different virtual stacking terms $ L_{\text{stack}}$, we improve model diversities by different training losses.

\begin{algorithm}[b]
\vspace{-1mm}
\caption{Adaptive Students Pruning}
\begin{algorithmic}[1]
\State \textbf{\#Adaptive prunning of student models}
\While{training}
	\State softLabel = TeacherModel(batch)		
	\For {$k \gets 1$ to StudentNum}
		\State rep=$\sum_j^{k}\alpha_j studentList[j](batch)$
		\State output = Classifer(rep)
		\State loss = softCrossEntropy(output,softLabel)
		\State Accumulate the gradients of current k students
	\EndFor
	\State Update all students based on the accumulated gradients
\EndWhile
\For {$k \gets 1$ to StudentNum}
		\State evaluate the previous $k$ students on validation set
\EndFor
\State Choose the best previous $n$ students
\end{algorithmic}
\label{algo:adaptive_pruning}
\vspace{-1mm}
\end{algorithm}

Besides the different training losses, \sys also brings in diversities from different perspectives.
All students have different initialized parameters.
The first student is initialized with a pre-trained language model with limited layers to improve the generalization, 
while any other student copies the parameters from the last trained student for the initialization (line 4 in Algorithm~\ref{algo:boosting}). 
Additionally, every student model is trained on different sampled subsets of training data (line 5), 
except only the first student is trained on the whole training data. 
For any student, we pick up the top $a\%$ samples with the largest residual errors between the teacher and the existing students. 
And we randomly sample another $b\%$ of all the training data. 
The subset of training data is augmented in the way suggested by previous works~\cite{jiao2019tinybert,hou2020dynabert}.

By following previous boosting ensemble theory~\cite{mason1999boosting,mason1999functional}, we make mathematical proof on the better convergence of the final representations of all students based on the gradient boosting ensemble as follows:
\begin{theorem}
    The MSE in \sys has the lower bound 0, 
    and it is a Lipschitz differentiable loss function (for any L > 1, we always have $|\bigtriangledown L_{\text{boost}}(x_1) - \bigtriangledown L_{\text{boost}}(x_1)| < L|x_1-x_2|$).
    Let $F^{(0)},F^{(1)},...$ be the sequence of combined hypotheses generated by the \sys training algorithm, using small enough step-sizes
    $\alpha_{i}:=-\frac{\langle \bigtriangledown L_{\text{boost}}(T, B^{(i-1)}),S^{(i)} \rangle} {L|S^{(i)}|^2}$.
    Then our sequential training of students either halts on round T with $- \langle \bigtriangledown L_{\text{boost}}(T, F^{(i-1)}),S^{(i)} \rangle \leq 0$, 
    or $L_{\text{boost}}(F^{(i)})$ converges to some finite value $L^*$, in which case
    \begin{equation}
        \lim_{i} \langle \bigtriangledown L_{\text{boost}}(T, B^{(t)}), S^{(t+1)} \rangle = 0 . 
    \end{equation}
\end{theorem} 
\noindent More proof details can be found in Appendix~\ref{sec:convergence_analysis}.

\subsection{Adaptive Student Pruning}
\label{sec:student_pruning}

By conducting adaptive student pruning,
we further train the classifier layer on top of the above final representation for better generalization and dynamical student numbers.
We follow the sub-network training in Network Architecture Searching (NAS)~\cite{cai2019once} to dynamically prune the redundant students for final representation extraction.
Then it trains the output classifier on the final representations extracted from different numbers of students to mimic the prediction output of the teacher.
Since the newly added students of boosting ensemble gradually improve accuracy by reducing the residual error, 
we drop and prune last-trained students without computing the importance scores in previous pruning works~\cite{hou2020dynabert}.
Because of Occam's razor theory,
student pruning can improve accuracy and generalization by removing the redundant students and only keeping the suitable group size. 
Moreover, adaptive student pruning enables dynamic student numbers with minimal accuracy loss.
For high throughput, we drop any numbers of the tailing students, while the remained students are fine-tuned in this setting to get higher accuracy. 
\S~\ref{sec:Hyperparameter} shows the better generalization of the best-pruned set of students and reduced accuracy loss of different numbers of students.

Specifically, we state the details about adaptive student pruning in Algorithm~\ref{algo:adaptive_pruning}. 
During the training of one data batch, 
It dynamically samples different numbers of students to train simultaneously (line 4 in Algorithm~\ref{algo:adaptive_pruning}). 
In this stage,  different numbers of ensembled students generate the final representation (line 5),
which has quite limited communication costs thanks to the small size of the pooled final representation. 
Then we put a new classifier layer on the sum of the selected students' outputs (line 6). 
We take the soft labels generated by the teacher model (line 3) as the target in the soft cross-entropy loss rather than the hard labels of the ground truth (line 7). 
Since there are multiple forward and back propagations of different numbers of students for the same training batch,
it accumulates all the gradients to update the parameters of the classifier and all students (line 8 and 10). 
Finally, we evaluate different numbers of previous $k$ students on the validation set to choose the optimal student number for the best-pruned students, instead of keeping all students (line 12-15).

\subsection{Online Inference}
\label{sec:inference}
\begin{figure}[t]
	\centering
		\includegraphics[width = 0.99\linewidth]{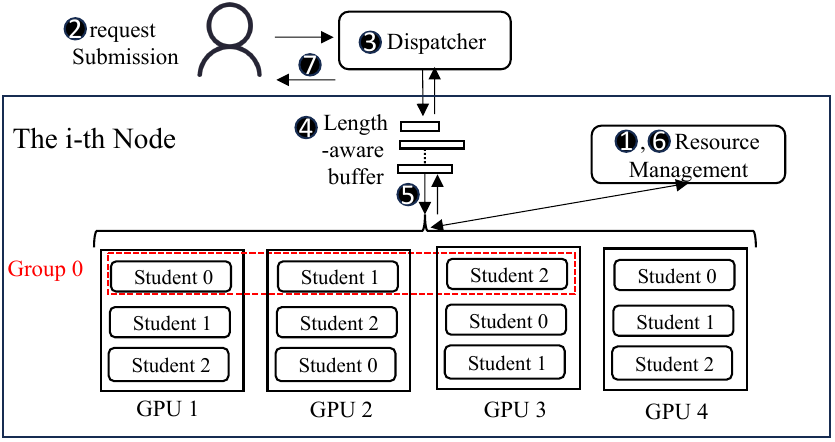}
		\caption{The online inference overview and procedure of student parallelism: student allocation and task dispatching
		 }	
		\label{fig:online_inference}
		\vspace{-8mm}
\end{figure}
For efficient GPU inference on online workloads,
we make specialized student allocation and inference sample dispatching for student parallelism.
Therefore, we can reduce the extra costs of waiting and padding to collect large and regular data batches for data parallelism.
We assume that the GPU cluster running online inference has multiple nodes,
and each node has multiple GPUs.
And we have one global dispatcher to balance the workloads among all nodes. 
Figure~\ref{fig:online_inference} shows the overview and procedure of online inference(more details in Appendix~\ref{sec:inference_overview}).

\subsubsection{Student Allocation}
\label{sec:model_allocation}

As shown in Figure~\ref{fig:online_inference}, 
\sys can distribute the different students over the multiple GPUs in the same node,
which can better parallelize the inference of every sample. 
Although the parallel students in the same group need communication to gather the final representation, 
it only takes about 0.2 ms due to the small size of the final pooled representation and fast NV-link or PCI-E.  
It further replicates the student group into multiple copies that are running on the same set of GPUs in the node.
Specifically, if the size of the student group is $S$ and total GPU number is $G$,
the $i$-th student of $j$-th student group is allocated as follows:
\begin{equation}
	\text{GPU}_{id} = (i+j \times S)\%G.
\end{equation} 
Therefore, if $i+j \times S > G$, the student co-locates on the same GPU with other students in the previous groups,
which run in parallel to process different individual samples concurrently.

During the inference, it keeps tracking the intensity of dynamic online workloads(i.e., the request per second) to adjust the student number accordingly. 
It temporarily decreases the student number (\S~\ref{sec:student_pruning}) to run more student groups for workload bursts, instead of reserving some idle GPUs in advance.
After workload bursts, \sys can increase the student number to fully utilize all GPUs for high accuracy.
More details about MPS and allocation hyper-parameters can be found in Appendix~\ref{sec:student_allcoation}).

\begin{table*}[h]
	\centering
	\small
	\begin{tabular}{lcccccccccc}
	\hline
	\multicolumn{1}{c}{}   &Param\#          & SST-2         & RTE           & MNLI          & CoLA(MMS)     & QQP           & MRPC(F1)      & STS-B(PCC) & QNLI          & Overall Score  \\ \hline
	*BERT-base-12L768D     & 109 M         & 92.7          & 67            & 83.6          & 52.8          & 89.6          & 88.6          & 89             & 91.8          & *81.89(100\%)        \\
 *I-BERT-12L768D        & 109 M         & 91.6          & 65.7          & 81.3            & 49.1          & 87.1          & 85.2          & 86.9              & 87.4          & *79.29(96.82\%)      \\
	DeeBERT-12L768D        & 109 M         & 91.5          & 66.1          & 80            & 43.4          & 87.3          & 85.2          & -              & 87.9          & 77.34(94.45\%)      \\
	DistillBERT-6L768D     & 67 M         & 91.3          & 58.4          & 81.1          & 49            & 88.1          & 86.9          & 86.9           & 88.9      &\underline{78.83(96.26\%)}\\
	DynaBERT-6L192D        & 9 M         & 92            & 63.2          & 82.15         & 43.7          & 90.4          & 81.4          & 87             & 88.5          & \underline{78.54(95.92\%)}    \\
	TinyBERT-4L312D        & 15 M         & 87.8          & 60.8          & 76.9          & 44.1          & 87.7          & 85.8          & 83.3           & 86            & 76.55(93.48\%)          \\
	PKD-BERT-3L768D        & 46 M         & 87.5          & 58.2          & 76.3          & 24.8          & 87.8          & 80.7          & 79.8           & 84.7          & 72.48(88.51\%)          \\
	BERT-2L256D            & 10 M         & 87.1          & 54.6          & 74.7          & 23.2          & 87            & 80.3          & 86             & 84.4          & 72.16(88.12\%)          \\
        \textbf{Academus-2L (pruned to 1)} & 10 M             & 91.3          & 64.9          & 77.3          & 42.3          & 88.1          & 86.7          & 85.2           & 88.2          & \underline{78.00(95.25\%)}    \\
	\textbf{Academus-2L (best pruned)} & 10*m M          & 91.8          & 65.9          & 78.2          & 43.2          & 88.7          & 86.9          & 86.4           & 89.8          & \textbf{78.86(96.31\%)} \\ \hline
	*BERT-large-24L1024D   &335 M  & 93.2          & 70.4          & 86.6          & 60.6          & 91.3          & 89.3          & 90             & 92.3          & 83.06(100\%)          \\
	\textbf{Academus-6L768D (best pruned)}   & 67*m M    & 92.1          & 61.4          & 82.5          & 51.2          & 89.2          & 88.9          & 87.2           & 89.7          & \textbf{80.28(96.65\%)} \\ \hline
	*DeBERTa-xlarge-48L1024D  &  658 M & 97            & 93.1          & 91.3          & 70            & 92.3          & 94.3          & 92.8           & 95.1          & 90.40(100\%)          \\
	Cocktail       & $\sum p_i$         & 94.2 & 72.1 & 87.1 & 62.1 & 91.5 & 89.7 & -     & 93.4 & 84.30(93.25\%) \\
	\textbf{Academus-12L768D (best pruned)} & 109*m M     & 94.1          & 78.7          & 88.8          & 59.3          & 91.9          & 90.1          & 91.5           & 93.1          & \textbf{85.94(95.06\%)} \\ \hline
	\end{tabular}
	\caption{Measurement and comparison on prediction quality: 
	* means the large model with full depth and width;
	xLxxxD means the model has x layers and xxx dimensions in the hidden state;
	the \textbf{bold} results means the better overall score than the other baselines without full depth or width; 
	and the \underline{underline} means the target quality achieved.
	}
	\label{tab:accuracy}
	\vspace{-6mm}
\end{table*}

\subsubsection{Inference Sample Dispatching}

\begin{figure}[b]
	\centering
		\includegraphics[width = 0.8\linewidth]{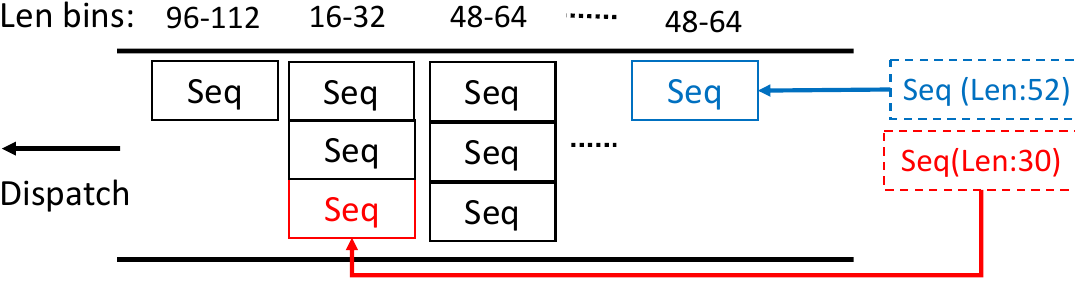}
		\caption{Length-aware Buffer: New sequence sample (e.g., the red one whose length is 30) can be merged with the previous element into a small batch, 
		since they share the same bin and the element size is not full. 
		Otherwise, the new sequence sample (e.g., the blue one of 52 tokens cannot join in the full element) is added to the buffer tail as a new element. }
		\label{fig:multi_queue}
		\vspace{-8mm}
\end{figure}
To reduce the extra costs of idle waiting and paddings for batching from online workloads,
\sys conduct the immediate and concurrent inference on different individual samples.
If there is any idle student group and no buffered samples,
the dispatcher can immediately send out the arrived sample for inference without waiting for batching.
Since different individual samples do not arrive at the same time, 
different student groups can concurrently process them by overlapping between data transferring to GPU and inference computation. 
Only if there are massive concurrent samples occupying all student groups,
the length-aware buffer temporarily stores the new samples, which merges the samples in the same length bin into small data batches with a few paddings.  
To efficiently merge samples on the fly, the length-aware buffer combines the queue with a dictionary whose key-value pair is length bin and queue position respectively to achieve $O(1)$ time complexity, as shown in Figure~\ref{fig:multi_queue}. 
More details about sample dispatching based on length-aware buffer can be found in Appendix~\ref{sec:dispatching}.

\subsubsection{Efficiency Analysis} We build a performance model to make a quantitative efficiency analysis. 
By considering all model and data factors like model depth $D$, model width $W$, batch size $B$, and input length $N$, waiting time $Q(B)$,
parallel computation capability $C$, the PCI-E transferring bandwidth $T$ , the GPU number $G$, and the total parallel model number $M$, we can model the latency as follows: 
\begin{equation}
    \text{Latency}= \Bigl\lceil \frac{WBN^2M}{CG} \Bigr\rceil + Q(B) + \frac{BN}{T} \\
\end{equation}
Obviously, \sys has the advantage of low model depth and is free of waiting and padding, compared to the existing works. Please refer to Appendix~\ref{sec:efficiency_analysis} for more details about the quantitative analysis and comparison.

\section{Evaluation}
\label{sec:evaluation}
In this section, we first introduce the experiment settings. 
Then we evaluate the overall performances of \sys. 
Finally, we conduct ablation experiments to study the influences of model hyperparameters and the sources of the improvement. 
From the evaluation, we can draw the following important conclusions:
\begin{itemize}
	\item \sys can effectively reduce more model depth while maintaining the best accuracy among small models without full depth or width. 
	Adaptive student pruning can achieve comparable accuracy with even only one student pruned from all students. 

	\item \sys can achieve about $ 4.1\times \sim 1.6\times$ speed up over all baselines in latency, 
	resulting from the smaller model depth and minimal waiting and paddings from data parallelism. 
	\item Based on adaptive student pruning, it can improve throughput up to $22.27\times$ compared to different methods.

\end{itemize}

\begin{table*}[t]
	\small
	\begin{tabular}{c|c|cc|cc|cc}
	\hline
	\multirow{2}{*}{Method} & \multirow{2}{*}{Model Config.(Param\#)} & \multicolumn{2}{c|}{Avg. Latency} & \multicolumn{2}{c|}{95\% Tail Latency} & \multicolumn{2}{c}{Throughput per GPU} \\ \cline{3-8} 
							&                                & Value (ms)    & Reduction (\%)    & Value (ms)    & Reduction (\%)    & Value(\#/s)     & Improvement ($\times$)  \\ \hline
	*BERT-base              & 12L768D (109 M)                      & 11.6          & 0.00\%            & 11.7          & 0.00\%            & 691.5           & 1.00                 \\
   *I-BERT        & 12L768D  (109 M)              & 4.9  & -57.8\% & 9.5 & -18.80\%  & 2115.99 & 3.06                 \\
        \textit{DeeBERT}        & \textit{12L768D  (109 M)}               & \textit{6.8}  & \textit{-41.38\%} & \textit{11.5} & \textit{-1.71\%}  & \textit{941.4} & \textit{1.36}                 \\
	DynaBERT                & 6L192D (9 M)                        & 8.3           & -28.45\%          & 9.6           & -17.95\%          & 7538.7          & 10.9                 \\
	Distilbert             & 6L768D (67 M)                         & 8.4           & -27.59\%          & 9.7           & -17.09\%          & 1508.1          & 2.18                 \\
	\textit{TinyBERT}       & \textit{4L312D (15 M)}                & \textit{7.3}  & \textit{-37.07\%} & \textit{9.7}  & \textit{-17.09\%} & \textit{5873.6} & \textit{8.49}                 \\
	\textit{PKD-BERT}       & \textit{3L768D (46 M)} 				 & \textit{6.8}  & \textit{-41.38\%} & \textit{9.5}  & \textit{-18.80\%} & \textit{2781.3} & \textit{4.02}                 \\
	\textit{BERT-2L}        & \textit{2L256D (10 M)}                & \textit{4.5}  & \textit{-61.21\%} & \textit{9.4}  & \textit{-19.66\%} & \textit{14420.3}& \textit{20.85}                \\
	\textbf{Academus-2L1S}    & 2L256D1S (10 M)                   & \textbf{2.8}  & \textbf{-75.86\%} & \textbf{3.6}  & \textbf{-69.23\%} & \textbf{15398.6}  & \textbf{22.27}           \\
    Academus-2L3S    & 2L256D3S  (30 M)                  & \textbf{3.0}  & \textbf{-74.13} & \textbf{3.7}  & \textbf{-68.37\%} & 520191  & 7.52           \\ \hline
	*BERT-Large             & 24L1024D (335 M)                      & 21.1          & 0.00\%            & 21.3          & 0.00\%            & 239.6           & 1.00                 \\
	\textbf{Academus-6L1S}	& 6L768D1S  (67 M)                  & \textbf{6.3}  & \textbf{-70.14\%} & \textbf{7.1}  & \textbf{-66.67\%} & \textbf{1508.1.5}    & \textbf{6.29}            \\
 \textbf{Academus-6L3S}	& 6L768D(1-3)S  (201 M)                  & \textbf{6.6}  & \textbf{-68.87\%} & \textbf{7.3}  & \textbf{-65.72\%} & 512.4    & 2.14            \\ \hline
	*DeBERTa-xlarge         & 48L1024D (658 M)                       & 40.5          & 0.00\%            & 41.4          & 0.00\%            & 160.9           & 1.00                 \\
	\textit{Cocktail}       & \textit{-}                     & \textit{21.4} & \textit{-47.16\%} & \textit{22.3} & \textit{-46.14\%} & \textit{240.4}  & \textit{1.49}         \\
	\textbf{Academus-12L1S}   & 12L1024D1S (109 M)                & \textbf{11.8} & \textbf{-70.86\%} & \textbf{12.6} &\textbf{-69.57\%}  & \textbf{621.5}    & \textbf{3.86}            \\
 \textbf{Academus-12L3S}   & 12L1024D3S (327 M)                & \textbf{12.1} & \textbf{-70.12\%} & \textbf{12.8} &\textbf{-69.08\%}  & 226.5    & 1.41            \\
 \hline
	\end{tabular}
	\caption{Measurement on inference efficiency: 
	* means the large model with full depth and width ;
	xLxxxD means the model has x layers and xxx hidden dimensions, and xS means the student number is x;
	the \textbf{bold} results means the better efficiency performance than the others without full depth or width;
	\textit{italic} indicates the method fails to satisfy the target accuracy.
	}
	\label{tab:efficiency}
	\vspace{-6mm}
\end{table*}

\subsection{Experiment Settings}
\subsubsection{Implementation Details}
In training,
we consider all 24 compact BERT variants~\cite{turc2019well} and all DEBERTA variants as the student candidates to find the shallowest one for different teachers to reach target accuracy. 
In practice, we use the BERT-2L256D~\cite{turc2019well} for BERT-base, BERT-6L768D for BERT-large, and Deberta-12L768D for DEBERTA-xlarge as the initialization of student models. 
 We conduct the same data augmentation with previous works~\cite{jiao2019tinybert,DBLP:conf/acl/XinTLYL20}. 
We conduct a grid search on various training hyperparameters to find the best for every dataset, 
including epoch number, batch size, learning rate, $\lambda$ weights, and data sampling ratio.
And in practice, we optimize the $\alpha_i$ together with the student, since they are both differentiable. 
We use 16 Geforce 3090 GPUs on 4 nodes and more implementation details can be found in Appendix~\ref{sec:impl_details}.

\subsubsection{Datasets and Metrics}
We follow previous works~\cite{DBLP:conf/acl/XinTLYL20,sanh2019distilbert,hou2020dynabert} to use the famous GLUE benchmark~\cite{wang2018glue}.
It consists of 8 text datasets in various text mining (SST-2, CoLA, MNLI, and RTE) and information retrieval (MRPC, QQP, STS-B, and QNLI) tasks.
We report various prediction quality metrics recommended by GLUE, including accuracy, F1 score, Matthews Correlation Coefficient (MCC), and Pearson correlation coefficient (PCC).
Additionally, we also build our workload generator upon the real production trace of twitter~\cite{khaleq2018cloud} to feed the inference system.
For text mining, we report the \textbf{average latency} and \textbf{the 95\% tail latency}.
We report \textbf{throughput} for information retrieval tasks to better show how long users will wait for the ranked results.

\subsubsection{Baselines}
We compare our \sys with various types of baselines related to BERT-like models. 
First, we consider the finetuned 12-layer \textbf{BERT-base-12L768D}~\cite{devlin2018bert}, 24-layer \textbf{BERT-large-24L1024D}~\cite{devlin2018bert}, 
and 48-layer \textbf{DeBERTa-xlarge-48L1024D}~\cite{he2021deberta} as the original teachers.
Then we conduct quantization to get \textbf{I-BERT-12L768D} with the same model architecture.
We train the 2-layer, 6-layer, and 12-layer versions of \sys for these teachers respectively, reducing over 75\% model depth.
And we make the \textbf{DeeBERT-12L768D}~\cite{DBLP:conf/acl/XinTLYL20} the baseline for the adaptive computing.
We use the 3-layer \textbf{PKD-BERT-3L768D}~\cite{sun2019patient}, 4-layer \textbf{TinyBERT-4L312D}~\cite{jiao2019tinybert}, and 6-layer \textbf{Distilbert-6L768D}~\cite{sanh2019distilbert} as the representatives of knowledge distillation. 
For pruning, we choose the fastest version of \textbf{DynaBERT-6L192D}~\cite{hou2020dynabert} satisfying the target prediction.
We also bring in extremely shallow finetuned models like \textbf{BERT-2L256D}~\cite{turc2019well} as the vanilla 2-layer model.
However, all these baseline works can only optimize the finetuned 12-layer BERT-base model.
We further compare our work with \textbf{Cocktail} based on bagging ensemble of all the above-mentioned models except our \sys 
to approach the performance of 48-layer DeBERTa-xlarge.
By following the previous works~\cite{zhang2019mark,gunasekaran2022cocktail}, all baselines employ the dynamical batch size for practical online workloads. 
We set the time budget as 10ms to dynamically set the batch size and waiting window according to online workloads~\cite{zhang2019mark}.

\subsection{Overall Performance}
In this subsection, we first verify if \sys can maintain the accuracy.
Then we show its advantages in average latency, tail latency and throughput. 

\subsubsection{Prediction Quality}
\label{sec:accuracy}

As shown in Table~\ref{tab:accuracy}, \sys can achieve the best overall score up to 96.32\% of BERT-base with only two layers, 
and it is the only one compressing BERT-large and DeBERTa-xlarge into their 25\% model depth while reaching the target accuracy. 
The results show other baselines can only reduce about half the layers of the BERT-base to maintain accuracy, 
which reach the target score\footnote{95\% overall score of the teachers, recommended by MLPerf~\cite{reddi2020mlperf}}.
Compared with tiny baselines with less than 6 layers, \sys has significant advantages in the prediction qualities.
Especially, the 3-layer PKD-BERT and 2-layer BERT-2L512D can only maintain less than 90\% overall score.
\sys is also better than any 6-layer or deeper reduced baselines with over 3$\times$ larger model depth.
It is even comparable with I-BERT always having full depth and width.
Because of improved generalization from adaptive student pruning,
our work can use even one student to handle workload burst and reach a competitive overall score (95.25\% of the BERT-base) compared with 6-layer baselines. 
If using deeper student models, \sys also can reduce the model depth of larger BERT-large and DeBERTa-xlarge by 75\%,
while the others fail.
Cocktail cannot catch up with DeBERTa-xlarge even with 24 layers. 
Because DeeBert and Cocktail need classification probability to work, they cannot handle regression dataset STS-B. 

\subsubsection{Latency}

In discriminative text mining tasks, 
like news sentiment classification for automatic trading, 
users only care about the inference latency of their own individual samples.
Therefore, Table~\ref{tab:efficiency} has two columns to show the average and tail latency of all methods in the SST-2 dataset. 
Our work has achieved both the lowest average and tail latency among all methods, 
significantly improving the others by $4.1\times\sim 1.6\times$.
The 3-student \sys is the best-pruned version, and it has slightly higher latency than the single student due to final communication among students. 
Compared with the others except for BERT-2L, \sys has a lower model depth to achieve low latency. 
Furthermore, it is free of long waiting time to collect enough data samples,
so it has a lower latency than BERT-2L. 
The different lengths are the main reason for small latency variances in \sys. 
Because average latency increases with the layer number of baselines, 
the expected computation latency of all models having over 6 layers has already been over the overall time budget. 
Other baselines like BERT-2L, PAK-BERT, TinyBERT, Distilbert, and DynaBERT are still below budget, 
but they have to wait longer to collect the data batch. 
I-BERT suffers from the original computation complexity and small batch sizes from online workload to get the higher latency. 
Because different samples exit from different layers, 
DeeBERT has the largest difference between the average and tail latency. 
The latency of Cocktail is determined by the largest model like 24-layer BERT-large to approach the accuracy of DeBERTa-xlarge. 
We omit similar results on the other datasets due to the limited space.

\subsubsection{Throughput}
In information retrieval tasks, like the search engine and QA system,  
the user has to wait to get the final result until dozens of document candidates are scored for the relevance of the query.
Hence, we report the throughput per GPU for best-pruned students and the single student to compare with other baselines on the same number of GPUs. 
Because some models have been out of memory, we set the batch size as 64 for all baselines.
The last column of Table~\ref{tab:efficiency} records the overall throughput / GPU number for all methods in the MRPC dataset.
Our work can improve throughput over the others by up to $ 22.27\times$, 
because it can leverage adaptive student pruning to use only one small student whose generalization has been improved (\S~\ref{sec:accuracy}). 
Meanwhile, even the best pruned of students (3 in total) can achieve comparable throughput with some competitive baselines like TinyBERT for better accuracy if the workload is not so heavy.
Except for the embedding layer, all model parameters mean computations, so the throughput negatively correlates with the model size. 
Even with fast integer calculation, the full model size of I-BERT limits its throughput improvement to $3\times$.
The 6-layer DynaBERT is the most competitive one of the small baselines reaching the target accuracy 
because it reduces model width to be $25\%$.
The vanilla BERT-2L256D has high throughout, but it is far away from the target accuracy.
Cocktail suffers from low throughput per GPU since it has to use multiple GPUs.

\subsection{Abalation Studies}
We further conduct ablation studies, including hyperparameters tuning, accuracy analysis, and latency analysis. 


\subsubsection{Model Hyperparameter Tuning}
\label{sec:Hyperparameter}
Since student number and model depth are key hyperparameters influencing the prediction quality, 
we conduct experiments to measure how they affect the prediction quality.
According to the results, we verify that adaptive student pruning can improve the generalization of the remained students and reduce the student number with minimal accuracy loss.
Furthermore, we can even further improve prediction quality with larger model depth with the relaxed latency requirement. 

\begin{figure}[b]
	\vspace{-6mm}
	\centering
	\subfigure[Student Number]{
		\begin{minipage}[b]{0.45\linewidth}
			\includegraphics[width=1\linewidth]{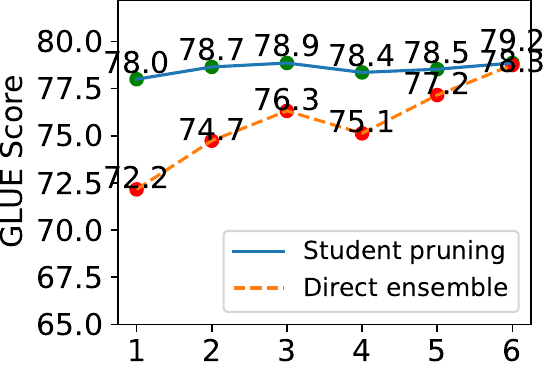} 
		\end{minipage}
		\label{fig:stu_num_acc}
	}
    \subfigure[Model Depth]{
    		\begin{minipage}[b]{0.45\linewidth}
   		 	\includegraphics[width=1\linewidth]{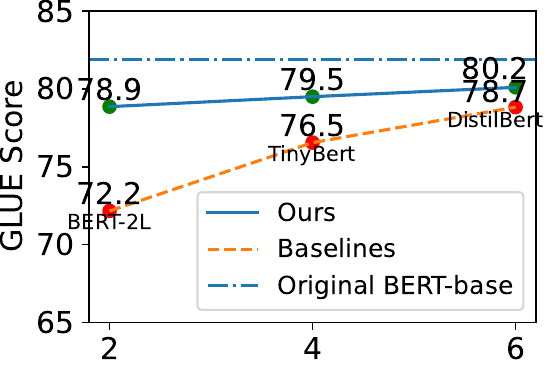}
    		\end{minipage}
		\label{fig:stud_depth_acc}
    	}
	\caption{Hyperparameter influences on prediction quality: (a) Different student number comparison. 
		(b) Different layer number influences comparison}
\end{figure}

\begin{table}[t]
	\centering
	\small
	\begin{tabular}{ccc}
	\cline{1-2}
						  & MRPC (F1) &                      \\ \cline{1-2}
	\sys             & 86.9       &                      \\ \cline{1-2}
	w/o pruning           & 86.8     &  	 				\\
	w/o stacking distillation & 86.0     &                      \\
	replacing boosting with bagging & 85.6     &                      \\ \cline{1-2}
	\end{tabular}
	\caption{Prediction quality ablation experiment results}
	\label{tab:quality_abalation}
    	\vspace{-6mm}
\end{table}
Figure~\ref{fig:stu_num_acc} shows that adaptive student pruning consistently outperforms the direct boosting ensemble of the same number of students without pruning in overall GLUE scores. 
The results verify that adaptive student pruning can effectively improve the generalization as a pruning technology. 
Because it improves the generalization of the remained students, 3 students are the best pruned to get the highest test performance in most datasets.
And even only one student pruned from all 6 students can maintain a relatively high score.
However, when we directly train the boosting ensemble of $m$ students without pruning, 
the limited total student number has more difficulties in finding the optimal solution.

As shown in Figure~\ref{fig:stud_depth_acc}, our \sys can further improve the prediction with a deeper student when learning from BERT-base in all datasets. 
The 4-layer students distilled from BERT-base can significantly improve the prediction results. 
\sys can even get a similar F1 score to the original 12-layer BERT-base if using 6-layer students.
And compared with other best-optimized baselines that have the same layers (such as BERT-2L256D, TinyBERT, and Distilbert), our \sys can consistently outperform them in overall GLUE scores,
while having the same layer number and similar inference latency. 

\subsubsection{Accuracy Analysis}

As shown in Table~\ref{tab:latency_abalation}, we first conduct accuracy ablation experiments to verify the effectiveness of our training methods to maintain accuracy on the MRPC dataset, as stated in Table~\ref{tab:quality_abalation}. 
When we directly use all trained students without adaptive student pruning, we observe that the F1 score decreases a little from 86.9 to 86.8. 
After removing the stacking distillation for better refinement, the F1 score further drops to 86.0, which is slightly lower than the TinyBERT-4L312D.
We further replace the boosting ensemble with a simple bagging ensemble of multiple small students, and the performance will further decrease.  
The same trend also happens on the other datasets.  

\subsubsection{Latency Analysis}
We also make breaking-down experiments to evaluate the influences and contributions of different design parts on inference latency.
The results show that latency increases dramatically when we increase the student's model depth to 4 layers, 
which indicates that reducing serial computation is the key to low latency.
After we pad all data samples into the maximum length, the average latency increases to be similar to tail latency. 
It means that the variances of \sys latency mainly come from the irregular sentence lengths. 
We further employ the waiting queue to batch enough samples for data parallelism, 
then average latency increases a lot and tail latency even gets close to the overall time budget.

\begin{table}[t]
\centering
\small
\begin{tabular}{ccc}
\hline
                      & avg (ms) & tail (ms) \\ \hline
\sys              	& 2.8             & 3.6               \\ \hline
4-layer students 	  & 4.5             & 5.2               \\
with padding         & 5.1             & 5.3               \\
with waiting queue    & 7.5             & 9.5              \\ \hline
\end{tabular}
\caption{Latency ablation experiment results}
\label{tab:latency_abalation}
\vspace{-6mm}
\end{table}

\section{Discussion}
\textit{Extention to Generative Models.}
The generative models generate tokens step by step, and every step makes a classification decision. Therefore, our work can be naturally extended to speed up every token classification. However, we do not conduct such experiments due to three main reasons:
(1) There is no golden standard (e.g., an objective metric like accuracy or F1 in discriminative tasks) to measure how well a sentence is generated (perplexity and BLEU are arguably only the indirect references instead of the golden one). It is relatively harder to verify if it gets good prediction quality. 
(2) We didn’t find the 2-layer pre-trained GPT like BERT-2L from well-read students~\cite{turc2019well} to initialize the generative students. And we cannot afford to collect such large datasets and conduct large-scale pre-training tasks. (3) Most importantly, we believe that speeding up BERT is non-trivial. They have been widely adopted by online services related to text mining and information retrieval. Information retrieval plays important roles like RAG~\cite{lewis2020retrieval} in the future because GPT suffers from the risks of hallucination.

\textit{Training Cost.} 
\sys has a similar time cost to other model compressions(e.g., 1.35x the training time of DynaBert with comparable accuracy), when using one GPU for training. Although we need to train multiple students, the student is a much smaller (e.g., 25\% depth and 50\% width) pre-trained model compared with other baselines. Each student is only trained on the sub-sampled dataset(e.g., 20\%) based on the residual error in the boost ensemble learning. Pruning and quantization also need extra training or calibration time.
Furthermore, the training of multiple students can be easily accelerated by multiple GPUs, without tuning parameters for larger batch sizes. Specifically, we can use one GPU to train the current student and distribute the inference of previous students on different GPUs.
Most importantly, online services like search engines can have much more inference samples (e.g., billions a day) than training samples, which can amortize the training costs.

\section{Conclusion and Future Work}
We propose \sys to adopt student parallelism to trade more students for less model depth. 
Every student model can have a decoupled and homogeneous architecture of fewer layers for low inference latency 
because all student models work together to maintain accuracy via boosting ensemble and stacking distillation.
And adaptive student pruning can temporally drop some students to improve the throughput for workload burst.
We further make specialized designs for student parallelism to minimize waiting and padding.
Comprehensive experiments with real online workloads have verified the effectiveness of our work in terms of latency, throughput, and accuracy.
In future work, we will extend it to generative models like GPT and reinforcement learning.

\bibliographystyle{plain}
\bibliography{references}

\newpage

\appendix
\section{Theory Analysis}
In this section, 
we first conduct the convergence analysis on our training methods based on boosting ensemble,
which shows the residual training loss converges to the infimum as more students added during the training. 
Then we make the quantitative analysis on the inference efficiency by comparing our proposed \sys and other typical baselines.
\subsection{Convergence Analysis}
\label{sec:convergence_analysis}
In this section, we conduct the convergence analysis on our training methods based on boosting ensemble,
which shows the residual training loss converges to the infimum as more students are added during the training. 
To prove the convergence, we can first prove our sequential training of students is one special case of Anyboost~\cite{mason1999boosting,mason1999functional},
then we can reuse the convergence properties and proofs of Anyboost in our proposed \sys.

Specifically, we take a non-parametric approach to view our sequential training of students as the numerical optimization problem in function space.
And we consider $B(x))$ evaluate at each point $x$ as a "parameter" 
and seek to minimize the following MSE error:
\begin{equation}
    L(T(x), B(x))) = \frac{1}{2}|T(x)-B(x))|^2 ,
\end{equation}
to make the ensemble of students generates a similar final representation with the original teacher $T(x)$.
In function space, there are an infinite number of such parameters, but in data sets only a finite number $\{B(x)_i)\}_1^N$ are involved. 
Following the numerical optimization paradigm we take the solution to be
\begin{equation}
    F^*(x)=\sum_{i=0}^M f^{(i)}(x) ,
\end{equation}
where $f^{(0)}(x)$ is the initial student $S^{(0)}(x)$ directly mimics the teacher, 
and any $f^{(i!=0)}(x)$ is the incremental function ("step" or "boost") defined by the optimization method as follows:
\begin{equation}
    f^{(i)}(x) = - \alpha_i g^{(i)}(x).
\end{equation}
And $g^{(i)}(x)$ is the gradient decent direction of $L(\cdot)$, namely: 
\begin{equation}
    \begin{split}
    g^{(i)}(x) = T(x)-B^{(i-1)}(x)
    \end{split}
\end{equation}
Therefore, our main loss $L_{\text{boost}}$  just makes the added i-th student $S^{(i)}(x)$ fit the $g_i(x)$:
\begin{equation}
    \begin{split}
    L_{\text{boost}}(x) = \frac{1}{2}\left\| T(x)-B^{(i-1)}(x)-S^{(i)}(x) \right\|^2 
        = \frac{1}{2}\left\| g^{(i)}(x)-S^{(i)}(x) \right\|^2 
    \end{split}
\end{equation} 

Although we apply other regularization like stacking distillation, 
the final representation of the student is only related with the above loss.    
Since the added i-th student $S^{(i)}(x)$ is trained to fit $g^{(i)}(x)$ in the final output and $\alpha_i$ is determined by the line serach,
the sequential training of students in \sys belongs to Anyboost~\cite{mason1999boosting,mason1999functional}.  
As a result, the boosting ensemble of students to mimic the teacher in \sys should share the same properties with Anyboost to have the following convergence theorem:
\begin{theorem}
    The MSE in \sys has the lower bound 0, 
    and it is Lipschitz differentiable loss function (for any L > 1, we always have $|\bigtriangledown L_{\text{boost}}(x_1) - \bigtriangledown L_{\text{boost}}(x_1)| < L|x_1-x_2|$).
    Let $F^{(0)},F^{(1)},...$ be the sequence of combined hypotheses generated by the \sys training algorithm, using small enough step-sizes
    $\alpha_{i}:=-\frac{\langle \bigtriangledown L_{\text{boost}}(T, B^{(i-1)}),S^{(i)} \rangle} {L|S^{(i)}|^2}$.
    Then our sequential training of students either halts on round T with $- \langle \bigtriangledown L_{\text{boost}}(T, F^{(i-1)}),S^{(i)} \rangle \leq 0$, 
    or $L_{\text{boost}}(F^{(i)})$ converges to some finite value $L^*$, in which case
    \begin{equation}
        \lim_{i} \langle \bigtriangledown L_{\text{boost}}(T, B^{(t)}), S^{(t+1)} \rangle = 0 . 
    \end{equation}
\end{theorem}
Because the detailed proof can be found in the previous theory work~\cite{mason1999boosting,mason1999functional}, we skip the proof for the convergence theorem.

\subsection{Efficiency Analysis for GPU Inference}
\label{sec:efficiency_analysis}

To make the quantitative analysis and comparison, we first build a performance model for the inference latency and throughput.
In this model, we consider all model and input factors, 
including model depth $D$, model width $W$, batch size $B$, and input length $N$.
Moreover, we also consider the time cost of the waiting queue $Q(B)$ that collect the maximum $B$ samples, 
and the parallel computation capability $C$, the PCI-E transferring bandwidth $T$ from the host to the GPU, the GPU number $G$, and the total parallel model or student group number $M$ on the cluster.
Therefore, we can model the inference latency of BERT-like models as follows: 
\begin{equation}
    \text{Latency}= \Bigl\lceil \frac{WBN^2M}{CG} \Bigr\rceil + Q(B) + \frac{BN}{T} \\
\end{equation}
The first term is the inference computation time on the GPU, 
in which $\Bigl\lceil \frac{WBN^2M}{CG} \Bigr\rceil$ is computation time of one layer and the ceiling symbol means it is constant if there is enough parallel computation capability $C$.   
And the second term $Q(B)$ stands for the waiting time of collecting the data batch, which depends on the workload and batch size B. 
The last term means the PCI-E transferring time of the data size $BM$.
Furthermore, we can further model the throughput per GPU based on the inference latency as follows:
\begin{equation}
    \text{Throughput Per GPU}= \frac{BM}{\text{Latency} \cdot G}
\end{equation}
Obviously, the total sample number $BM$, the GPU number $G$, and the latency determine the throughput per GPU together.

\begin{table}[b]
    \resizebox{\linewidth}{!}{
    \begin{tabular}{cccccccc}
    \hline
              &  Depth    &  Width        & Batch Size      &  Length            & M                        &  Q  \\ \hline
    BERT-base & 12        & 768           & $\sim$10        & 128                & G                   & Yes                     \\
    TinyBERT  & 4         & 312           & $\sim$10        & 128                & G                   & Yes                     \\
    DynaBERT  & 6         & 192           & $\sim$10        & 128                & G                  & Yes                     \\
    DeeBERT   & 1-12      & 768           & $\sim$10        & 128                & G                   & Yes                     \\
    Cocktail  & $\max(D_i)$ & $\sum_i W_i$ &$\sim$10        & 128                & G                   & Yes                     \\ \hline
    \textbf{\sys}  & 2         & (1-3)*256     &  $\leq4$    & 8(n\%8+1)     & 4G                  & None          \\\hline  
    \end{tabular}
    }
    \caption{The comparison on all related factors with latency and throughput:
     M is the number of running model or student group, G is the GPU number, and Q is the waiting queue.}
    \label{tab:factors}
    \vspace{-5mm}
\end{table}

To analyze our advantages in efficiency,
we then make quantitative comparisons between \sys and some representative baselines in optimizing the BERT-base on the MRPC dataset.
As all the related factors shown in Table~\ref{tab:factors},  \sys can outperform the other baselines due to its model architecture and system designs.
Specifically, \sys can achieve the lowest model depth of only 2 layers to significantly decreases the latency.
Although the overall model width of 3 students brings in more parallel operators, 
the GPUs have enough parallel computation capability to speed them up.
Because \sys realizes the direct inference on different individual samples, there is no waiting $Q$ and large batch $B$ for PCI-E transferring.
And the single sample and small batches of 2$\sim$4 samples in the similar length has the input length as the closest bin length 8(n\%8+1).
In terms of throughput, the multiple students sharing the same GPU make the $M$ can be several times larger than $G$, 
leading to large enough total concurrent sample number $BM$ to fully utilize the GPUs.
Additionally, adaptive student pruning can reduce the student number as low as only 1, 
which can improve the student group number $M$ to achieve the highest thought.

In contrast, the other baselines suffer from the model architecture unfriendly to GPUs and extra costs for data parallelism.
The knowledge distillation baseline TinyBERT has 4 layers fewer than any other baselines,
but it is 2 times larger than \sys.
In adaptive computing baseline DeeBERT, some "easy" sample can exit from the bottom layers, 
while other "hard" ones still have to go throughpu all 12 layers.
Because Cocktail is the bagging ensemble of different models, 
its model depth is determined by the straggler that have the largest depth max($D_i$). 
Although the pruning baseline DynaBERT have the smallest model width of 192 hidden dimensions in all methods,
it mainly reduces the parallel operators that can be efficiently accelerated by GPUs.
And all baselines need the large enough data batches for data parallelism, 
brining in considerable waiting queue costs $Q$ and large PCI-E transferring time for the whole batch.
Traditionally, they have to pad all short samples to have the same maximum length $N$ as 128.
Otherwise, they need to suffer from extra compiling cost or computation divergence for the ragged batch.
Based on data parallelism, all the baselines only run one model on every GPU to process different data batches respectively, 
resulting in $M=G$.

\section{Implementation Details}
\label{sec:impl_details}
\subsection{Additonal Experiment Settings}
\label{sec:additonal_setting}
By default in Fig. 1, the model is a 12-layer BERT-base with 768 hidden dimensions, and the inference data batch consists of 8 samples with length of 256. In every sub-figure, we choose one as the changing variable and fix the others to measure the latency and throughput and show the influences of the changing variable.

\subsection{Testbed and Environment}
We use 4 servers each with 4 NVIDIA Geforce 3090 GPUs, Intel Xeon(R) E5-268340 CPU, and 96 GB memory for all evaluations. 
The server is supported by a Network File System (NFS) to store the models and datasets in the storage node. 
And it runs on the Ubuntu 18.04 operating system. 
Except the global dispatcher, every server has its own logical controller and computation workers.
And we implement the \sys with Python. 
Specifically, we use the PyTorch~\cite{pytorch2018pytorch} as our DL framework, 
HuggingFace Transformers~\cite{wolf2019huggingface} for the model implementations, 
, NCCL~\cite{jeaugey2017nccl} for the communications among students.

\subsection{Training Implementation Details}
In the offline training,
we consider all 24 compact BERT variants~\cite{turc2019well} and all DEBERTA variants as the student candidates to find the lowest one for different teachers. 
In practice, we use the BERT-2L256D~\cite{turc2019well} for BERT-base, BERT-6L768D for BERT-large, and Deberta-12L768D for DEBERTA-xlarge as the backbone network and initialization of student models respectively. 
Following previous works~\cite{jiao2019tinybert,DBLP:conf/acl/XinTLYL20}, we conduct the same data augmentation on the down-stream datasets. 
We conduct a grid search on various training hyperparameters to find the best for every dataset, 
including epoch number, batch size, learning rate, $\lambda$ weights, and data sampling ratio.
And in practice, we optimize the $\alpha_i$ together with the student, since they are both differentiable. 
 The training epoch number is early stopped by the validation loss. lr=5e-5 and betas=(0.9, 0.999) for Adam optimizer, $\lambda=1$ works fine in most cases, data sampling ratio is usually 0.2 (smaller in some large datasets).

The student number is determined by the validation loss. When conducting student distillation, we continue to add new students until the validation loss does not decrease any more. Then we conduct adaptive student pruning and measure the validation loss to decide the number of students to maintain. To determine the student model depth, we conduct the binary search from one layer to 25\% layers of the teacher model. We choose the smallest model depth that can reach the target accuracy threshold (i.e., 95\% of the teacher’s accuracy)


\subsection{Online Inference Implementation}
\label{sec:inference_overview}
For online inference, the global dispatcher to assign samples among different nodes in the round-robin method. 
And every node employs message queues to dispatch samples and collect the results in the master-worker architecture.
The master process is responsible for logical controlling, including resource management, the length-aware buffer, and task dispatching. 
And workers host all concurrent student groups on all the GPUs.
Once a user submits an inference request, the master preprocesses, buffers, and then dispatches the inference samples. 
And any idle worker can fetch the inference task from the message queue and return the results.
We employ the Multi-Process Service (MPS)~\cite{mps} to make multiple students of different groups share the same GPUs without interference.
As shown in Fig~\ref{fig:online_inference}\, sys has the following procedures during the online inference phase:
\begin{enumerate}[leftmargin=10pt]
\item In the initialization, the resource manager starts multiple student groups running in parallel, 
and every group distributes its students over multiple GPUs.
\item  User submits a request triggering one or more inference samples.
\item The dispatcher assigns the samples to different GPU nodes in the round-robin manner.
\item After reaching the GPU node, the sample is pre-processed and then enters the length-aware buffer. And if the buffer is not empty, it tries to merge the samples of the similar length into small batches.
\item Whenever any student group is idle, it immediately sends out the first buffered data sample or batch of samples in similar lengths for immediate inference without waiting. 
\item According to the dynamic intensity of online workload, resource management decreases the student number for larger throughput or increases it for better accuracy. 
\item It returns results to users. 
\end{enumerate}

\begin{figure}[t]
\centering
	\includegraphics[width = 0.99\linewidth]{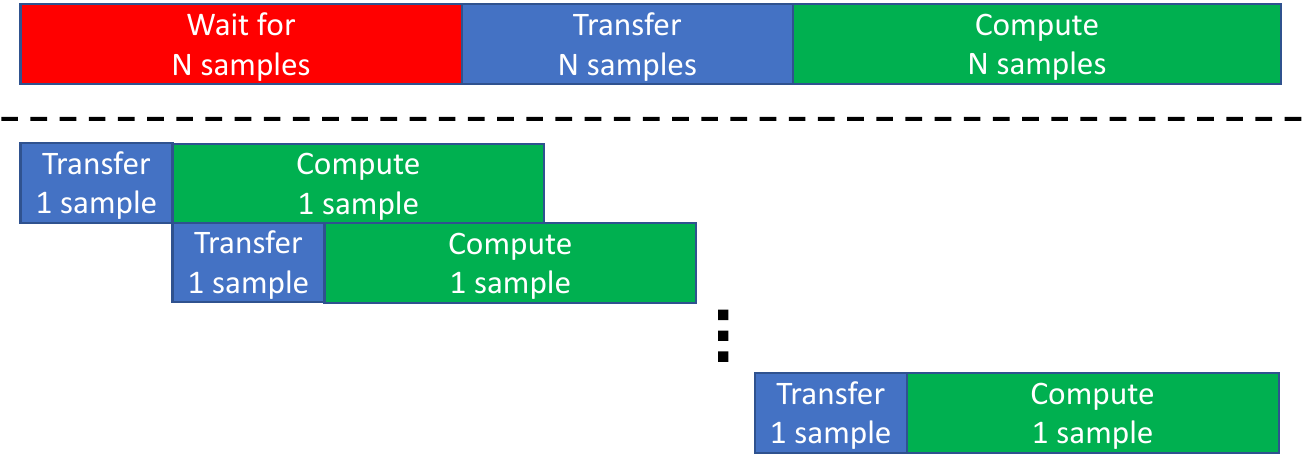}
	\caption{ Batch VS. Concurrent Samples: the top is the timeline of batching and processing all samples without overlapping, 
	and the bottom is how multiple models concurrently process different samples without batching to overlap the data transferring and inference computations.
	 }	
\label{fig:overlapping}
 \vspace{-5mm}
\end{figure}
\subsubsection{Details about Student Allocation}
\label{sec:student_allcoation}
\sys employs NVIDIA Multi-Process Service (MPS)~\cite{mps} to make multiple students from different groups share the same GPU without interference. 
It can guarantee different students on the same GPU have separate computation units and memory bandwidth to run concurrently.
Since different individual samples do not arrive at the same time, 
different student groups can concurrently process them by overlapping between data transferring to GPU and inference computation as shown in Figure~\ref{fig:overlapping}. 

The resource management determines the proper hyperparameters of student allocation. 
In the initialization, it conducts profiling and grid searching to find the optimal student number sharing the same GPU and the maximum batch size of each student group (used in \S~\ref{sec:dispatching}).
Both hyperparameters are set to maximize the total throughput without increasing latency due to lacking computation capability.
During the inference, it dynamically adjusts the student number to host different numbers of student groups according to the workload intensity and resource usage. 
It keeps tracking the buffer size (decribled in \S~\ref{sec:dispatching}) and the idle group number as indicators of workload intensity and resource usage. 
If the buffer size reaches the threshold (i.e. the total student group number), 
\sys drops one last-trained student models to start more groups, so that the waiting time of buffered samples will not exceed the inference time. 
And if the buffer is empty and the total student number of all idle groups is larger than occupied group number for a certain time window (e.g., 2 minutes),
it means there are enough free resources to increase the student number by one for all the occupied student groups. 
Then \sys can make better use of idle computation resources to improve accuracy.

\subsubsection{Details about Sample Dispatching based on Length-aware Buffer}
\label{sec:dispatching}
Although student parallelism reduces the necessity of data parallelism via direct inference on individual samples, 
we further propose the length-aware buffer to merge the concurrent samples of similar lengths into small batches,
in case that all student groups are occupied.
Different from the waiting queue that keeps waiting until enough samples arrive,
the length-aware buffer only temporarily stores the new samples if all student groups have been occupied by previous samples.
If idle student group is available and the length-aware buffer is empty, \sys directly sends out the new sample for the direct inference.
Once the inference sample arrives, we check which length bin it belongs to and pad or clip it into the maximum length of the bin.
Only if the buffer is not empty, it tries to merge the new sample and others in the same length bin into a small data batch as one buffer element on the fly.
Whenever any running group is idle, the dispatcher immediately sends out the first buffer element (a sample or small batch) for inference without any delay.
We set the maximum buffer size as the total number of current student groups, 
so the range of buffer time is from zero to the inference latency.

Figure~\ref{fig:multi_queue} shows the details about how the length-aware buffer efficiently generates small batches of samples in similar lengths on the fly.
We equally split the full length range (i.e., 0-128) into 16 bins at every step of 8 tokens. 
Then we set the bins and pointer as the key-value pair in the dictionary to quickly index samples of different length bins. 
When a new sample arrives, we check the dictionary to see if any sample belongs to the same length bin in the buffer. 
If so and the element size is not the maximum (e.g., 4, which is also determined by profiling), we can get the element pointer from the dictionary to merge them as one buffer element. 
Otherwise, we append the new sample as the tail element in the buffer and update its length bin and pointer in the dictionary. 
When the first buffer element is dispatched, we delete its key-value pair in the dictionary.
Obviously, all the operators of the length-aware buffer only has $O(1)$ time complexity due to the hash map. 

\subsubsection{Student Number Adaptation} 
If there are intensive workloads to fulfill the length-aware buffer, 
\sys can further decrease the student number (\S~\ref{sec:student_pruning}) to run more student groups.
According to the workload intensity and resource usage, 
the resource manager dynamically adjust the student number to host different numbers of student groups on the available GPUs.
It keeps tracking the buffer size and the idle group number as indicators of workload intensity and resource usage. 
If the buffer size reaches the threshold (i.e. the total student group number), 
\sys drops one last-trained student models to start more groups, so that the waiting time of buffered samples will not exceed the inference time. 
And if the length-aware buffer is empty for a certain time window (e.g., 2 minutes), 
\sys checks whether total student number of all idle groups is larger than occupied group number.
If so, it means there are enough free resources to increase the student number by one for all the occupied student groups. 
Then \sys can make better use of computation resources of the idle students to improve accuracy.
To avoid cold starting for the entire environment, 
the inactivated student instances of different student number settings can be preloaded in the host memory to realize hot standby.

\end{document}